%% file: main.tex
\documentclass[letterpaper]{article} 
\usepackage[preprint]{aaai2027}  
\usepackage[hyphens]{url}  
\usepackage{graphicx} 
\usepackage{natbib}  
\usepackage{caption} 
\usepackage{algorithm}
\usepackage{algorithmic}

\usepackage{newfloat}
\usepackage{listings}
\DeclareCaptionStyle{ruled}{labelfont=normalfont,labelsep=colon,strut=off} 
\floatstyle{ruled}
\newfloat{listing}{tb}{lst}{}
\floatname{listing}{Listing}

\usepackage{booktabs}

\usepackage{amsmath,amsfonts,bm}
\usepackage{algorithm,amsmath,amssymb,bm,amsthm}
\usepackage{algorithmic}
\usepackage{enumitem}

\newtheorem{lemma*}{Lemma}

\title{Respect Your Zero-Shot Uncertainty: Conservative Calibration for Test-Time-Adapted Vision-Language Models}
\author{
    Jingyan Jiang\textsuperscript{\rm 1}\equalcontrib,
    Yaru Sun\textsuperscript{\rm 1,\rm 3}\equalcontrib,
    Xiao Chen\textsuperscript{\rm 2},
    Jiazhen Huang\textsuperscript{\rm 2},
    Caiting Li\textsuperscript{\rm 1},
    Zhijian He\textsuperscript{\rm 1},
    Yin Chen\textsuperscript{\rm 1}\corresponding,
    Pingting Hao\textsuperscript{\rm 4}\corresponding
}

\affiliations{
    \textsuperscript{\rm 1}Shenzhen Technology University,
    \textsuperscript{\rm 2}Tsinghua University,
    \textsuperscript{\rm 3}Shenzhen University,
    \textsuperscript{\rm 4}Northeast Normal University

    jiangjingyan@sztu.edu.cn,
    2510263033@mails.szu.edu.cn
}

\begin{document}

\maketitle



\input{sections/0-abstract}

\input{sections/1-intro}
\input{sections/2-related}

\input{sections/3-observation1}
\input{sections/4-method1}
\input{sections/5-experiment}
\input{sections/6-conclusion}

\bibliography{aaai2027} 
\end{document}

%% file: sections/0-abstract.tex
\begin{abstract}
Test-time adaptation (TTA) can improve the recognition accuracy of vision-language models under distribution shift, but often degrades calibration, making predictive confidence unreliable for downstream decision-making. Many existing label-free calibration approaches are either coupled to prompt optimization or rely on logit-range statistics that provide only a coarse characterization of the predictive distribution. We show that TTA can increase confidence and reduce entropy even when the top-1 prediction and its correctness remain unchanged, a failure mode we term prediction-preserving sharpening. Across diverse TTA methods and benchmarks, larger entropy reductions relative to paired zero-shot predictions are associated with greater increases in Expected Calibration Error (ECE). On entropy-reduced samples, confidence gains also tend to exceed accuracy gains. Based on these findings, we propose \textbf{Zero-Shot-Anchored Entropy Calibration (ZAEC)}, a label-free post-hoc method that uses zero-shot entropy as a sample-specific uncertainty reference. ZAEC selectively restores the zero-shot entropy of sharpened predictions through minimal temperature scaling while leaving all other predictions unchanged. It requires no labeled calibration data or learned parameters and preserves class rankings and classification accuracy. Across five TTA methods and 15 datasets, ZAEC achieves the lowest post-hoc macro-average ECE on ViT-B/16, with consistent gains on RN50.
\end{abstract}

%% file: sections/1-intro.tex
\section{Introduction}

Test-time adaptation (TTA) provides a practical way to adapt vision-language models (VLMs) to distribution shifts using only unlabeled target samples at inference time~\cite{TTA_survey,TTA_survey_2,Tent,chen2026neural,tpt,tda}. Yet improved recognition accuracy does not necessarily imply more reliable predictions. Recent evidence shows that many VLM TTA methods substantially degrade model calibration, causing predictive confidence to deviate from actual correctness~\cite{vlmtta}. Such miscalibration makes it difficult to determine when adapted predictions should be trusted and may produce high-confidence errors that mislead downstream decision-making. This issue is particularly concerning in safety-critical applications such as healthcare~\cite{health1} and autonomous driving~\cite{autonomous-driving1,autonomous-driving2}. Therefore, effective VLM adaptation should improve recognition performance without sacrificing the reliability of predictive confidence.

\begin{figure}[t]
    \centering
    \includegraphics[width=\columnwidth]{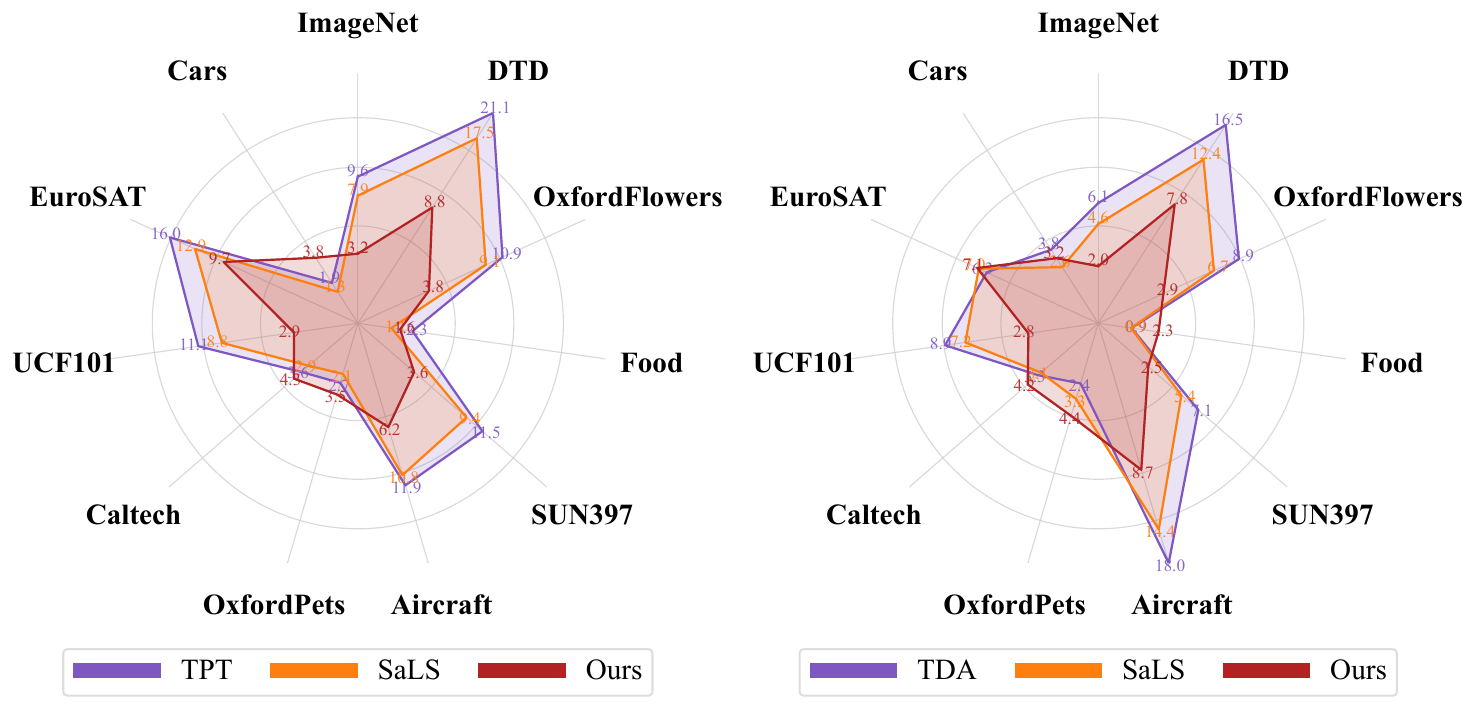}
    \caption{Expected Calibration Error (ECE) of raw TTA, SaLS, and ZAEC across 11 datasets for TPT and TDA. ZAEC achieves the lowest average ECE in both settings. \textbf{Lower is better.}}
    \label{fig:intro}
        \vspace{-0.2cm}
\end{figure}

Existing label-free calibration methods for VLM TTA mainly follow two directions. Prompt-coupled methods incorporate calibration objectives into test-time prompt optimization by encouraging dispersion, orthogonality, or angular diversity among class-wise text features~\cite{ctpt,otpt,atpt}. Their dependence on learnable prompts and text-feature geometry limits their applicability to cache-, prototype-, and logit-based adaptation. Plug-in methods are more general because they operate directly on adapted outputs. SaLS~\cite{sals}, for example, rescales each adapted logit vector to match the range of its zero-shot counterpart. However, the logit range depends only on the largest and smallest values and therefore ignores how probability mass is distributed among the remaining classes. A general post-hoc solution calls for a label-free, sample-specific signal that captures changes in the complete predictive distribution.

To identify such a signal, we examine how TTA reshapes predictive distributions relative to their paired zero-shot outputs. Our analysis reveals a consistent failure mode, which we term \textit{prediction-preserving sharpening}: TTA can increase confidence and reduce entropy even when the top-1 prediction---and hence its correctness---remains unchanged. More broadly, across diverse TTA methods and benchmarks, larger entropy reductions relative to paired zero-shot predictions are associated with larger increases in calibration error. On entropy-reduced samples, confidence gains also tend to outpace accuracy gains, indicating that adaptation can sharpen the predictive distribution beyond the observed improvement in correctness. These findings establish \textit{zero-shot-relative entropy reduction} as a label-free signal of potentially unsupported predictive sharpening. Moreover, the paired zero-shot entropy records each sample's uncertainty before adaptation, providing a conservative, sample-specific reference for measuring the additional concentration introduced by TTA.

Motivated by these findings, we propose \textbf{Zero-Shot-Anchored Entropy Calibration (ZAEC)}, a label-free post-hoc method that selectively corrects TTA-induced predictive sharpening. For samples whose adapted entropy falls below the zero-shot reference, ZAEC applies the smallest temperature required to restore it; all other predictions remain unchanged. Positive temperature scaling preserves class ordering, allowing ZAEC to retain the complete adapted ranking, top-1 prediction, and classification accuracy. Using only paired zero-shot and adapted logits, ZAEC requires neither labels nor learned parameters and supports diverse TTA mechanisms. Across five representative methods and two benchmark groups, ZAEC consistently achieves the lowest method-level average ECE among the evaluated post-hoc methods. Our main contributions are summarized as follows:

\begin{itemize}
    \item We reveal \textbf{prediction-preserving sharpening} as an important source of TTA-induced calibration degradation and show that zero-shot-relative entropy reduction is strongly associated with increased calibration error across methods and benchmarks.
    
    \item We introduce \textbf{ZAEC}, a label-free and mechanism-agnostic post-hoc method that selectively restores the zero-shot entropy of sharpened predictions through sample-wise temperature scaling. ZAEC requires no learned calibration parameters and provably preserves class rankings and classification accuracy.

    \item  Across five TTA methods and 15 datasets, ZAEC achieves the lowest method-level average ECE among the evaluated post-hoc methods on ViT-B/16, with consistent gains on RN50.
    
\end{itemize}

%% file: sections/2-related.tex
\section{Related Work}
\subsection{Test-time Adaptation for Vision-Language Models}

Test-time adaptation updates a pre-trained model or its prediction process using unlabeled target data during inference~\cite{Tent,note,cliff,fan2026moetta,chen2026neural,ttd}. Recent VLM TTA methods adapt different components of the CLIP inference pipeline. Prompt-based approaches optimize textual context tokens or prompt-conditioned representations for each test input~\cite{tpt,difftpt}. Prototype-based~\cite{tps,dpe,dlae}  approaches adjust text-derived class prototypes or jointly evolve visual and textual prototypes. Cache-based~\cite{tda,dmn} approaches accumulate reliable test features and pseudo-label information to refine subsequent predictions. These methods employ different adaptation mechanisms but
primarily focus on recognition accuracy. A recent systematic
benchmark shows that many VLM TTA methods generally degrade the calibration of the original CLIP model~\cite{vlmtta}.

\subsection{Calibration of Vision-Language Models}
Model calibration measures the agreement between predictive confidence and empirical correctness, with Expected Calibration Error (ECE) commonly used as an evaluation metric~\cite{guo2017calibration,naeini2015obtaining}. Label-free calibration methods for VLM test-time adaptation can be divided into adaptation-coupled and post-hoc approaches.
Adaptation-coupled methods integrate calibration objectives into test-time prompt optimization. C-TPT~\cite{ctpt} promotes dispersion among class-wise text features, O-TPT~\cite{otpt} imposes orthogonality constraints, A-TPT~\cite{atpt} maximizes angular diversity, and SoC~\cite{soc} preserves semantic structure while separating text features. These methods are designed around prompt-dependent text representations.
Post-hoc methods calibrate the final adapted logits and remain independent of the internal adaptation mechanism. SaLS~\cite{sals} uses the zero-shot logit range to determine a sample-specific scaling factor. ZAEC also uses the paired zero-shot prediction, while defining the correction through predictive entropy and applying it only to samples whose distributions become sharper after adaptation. The method is compatible with various types of test-time adaptation methods.

%% file: sections/3-observation1.tex
\section{Rethinking VLM Calibration}
\label{sec:motivation}

This section examines how test-time adaptation changes predictive uncertainty and how these changes relate to calibration degradation. We first establish the notation, then isolate predictive sharpening on samples whose top-1 predictions remain unchanged. We next evaluate the limits of logit-range statistics and analyze calibration degradation through entropy change relative to the paired zero-shot prediction.

\subsection{Preliminaries}
\label{sec:analysis_preliminaries}

Let $\mathcal{C}=\{c_1,\ldots,c_K\}$ denote the candidate class set. For a test sample $x_i$, the frozen zero-shot VLM and a test-time adaptation method produce logits $\mathbf{z}_i^{\mathrm{ZS}}\in\mathbb{R}^{K}$ and $\mathbf{z}_i^{\mathrm{TTA}}\in\mathbb{R}^{K}$, respectively. Their predictive distributions are
\begin{equation}
\mathbf{p}_i^{\mathrm{ZS}}
=
\operatorname{softmax}\left(\mathbf{z}_i^{\mathrm{ZS}}\right),
\qquad
\mathbf{p}_i^{\mathrm{TTA}}
=
\operatorname{softmax}\left(\mathbf{z}_i^{\mathrm{TTA}}\right).
\label{eq:analysis_prediction_distributions}
\end{equation}
For $s\in\{\mathrm{ZS},\mathrm{TTA}\}$, the predicted class and confidence are
\begin{equation}
\widehat{y}_i^{s}=\arg\max_k p_{i,k}^{s},
\qquad
c_i^{s}=\max_k p_{i,k}^{s}.
\label{eq:analysis_prediction_confidence}
\end{equation}
For a predictive distribution $\mathbf{p}\in\Delta^{K-1}$, its entropy is
\begin{equation}
H(\mathbf{p})
=
-\sum_{k=1}^{K}p_k\log p_k.
\label{eq:analysis_predictive_entropy}
\end{equation}
Lower entropy indicates a more concentrated predictive distribution. Ground-truth labels are used only for evaluation and the diagnostic analyses in this section. 

\textbf{Calibration Metric.}
Calibration is measured using the top-label Expected Calibration Error (ECE). We partition predictions into \(M\) confidence bins, where
\(B_m\) denotes the set of samples assigned to the \(m\)-th
bin, \(N\) is the total number of samples, and
\(\operatorname{acc}(B_m)\) and \(\operatorname{conf}(B_m)\)
denote the empirical accuracy and mean confidence within
that bin, respectively. ECE is computed as
\begin{equation}
\operatorname{ECE}
=
\sum_{m=1}^{M}
\frac{|B_m|}{N}
\left|
\operatorname{acc}(B_m)
-
\operatorname{conf}(B_m)
\right|.
\label{eq:ece}
\end{equation}
A lower ECE indicates better agreement between predictive
confidence and empirical accuracy.
\subsection{TTA-Induced Predictive Sharpening}
\label{sec:predictive_sharpening}
\paragraph{Prediction-preserving sharpening.}
We first separate predictive sharpening from changes in the predicted class. The prediction-agreement (PA) subset contains samples satisfying $\widehat{y}_i^{\mathrm{ZS}}=\widehat{y}_i^{\mathrm{TTA}}$, and the prediction-disagreement (PD) subset contains the remaining samples. Within the PA subset, the predicted class and sample correctness remain unchanged, which allows confidence and entropy changes to be examined without the confounding effect of label switches.

\textbf{Observation 1: Prediction-Preserving TTA Sharpens the Predictive Distribution.}
Table~\ref{obs1:table} shows that PA samples account for 61.21--89.93\% of the ImageNet-A test set across the evaluated methods. On these samples, TTA increases mean confidence by 8.78--29.74 percentage points, reduces entropy by 0.347--1.279, and increases ECE by 8.59--20.34 points. Accuracy remains fixed by the PA definition. TTA can thus produce substantial calibration degradation through predictive sharpening even when it preserves the top-1 prediction. Results for the PD subset are reported in Appendix~C.

\begin{table}[t]
\centering
\begin{tabular}{lrrrr}
  \toprule
  Method & Ratio (\%) & $\Delta$Conf. & $\Delta H$ & $\Delta$ECE \\
  \midrule
  TPT & 82.85 & +12.90 & -0.498 & +10.50 \\
  TPS & 65.89 & +29.74 & -1.279 & +20.34 \\
  TDA & 89.93 &  +8.78 & -0.347 &  +8.59 \\
  DPE & 61.21 & +29.48 & -1.209 & +19.18 \\
  \bottomrule
\end{tabular}
\caption{\textbf{Prediction-preserving samples on ImageNet-A with CLIP ViT-B/16.} All changes are computed as raw TTA minus zero-shot CLIP within the PA subset. Accuracy is unchanged by construction. The complete PA/PD analysis is provided in Appendix~C.}
\label{obs1:table}
\end{table}

\paragraph{Why logit-range statistics are insufficient.}
Existing post-hoc calibration methods such as SaLS use the paired zero-shot logit range as a sample-wise reference. For logits $\mathbf{z}_i$, the range is
\begin{equation}
R(\mathbf{z}_i)
=
\max_k z_{i,k}-\min_k z_{i,k}.
\label{eq:logit_range}
\end{equation}
This statistic depends only on the largest and smallest logits. It does not capture how probability mass is distributed across the remaining classes. For method $m$, we measure the relative deviation from the zero-shot range as
\begin{equation}
\Delta_R^{m}
=
\frac{1}{N}
\sum_{i=1}^{N}
\frac{
\left|R(\mathbf{z}_i^{m})-R(\mathbf{z}_i^{\mathrm{ZS}})\right|
}{
R(\mathbf{z}_i^{\mathrm{ZS}})
}
\times 100.
\label{eq:range_deviation}
\end{equation}
Figure~\ref{fig2}(a) compares $\Delta_R^{m}$ with ECE across four calibration approaches. TPT+SaLS nearly eliminates the range deviation and retains relatively high ECE. TPT+ZAEC achieves lower ECE with a nonzero range deviation on the same underlying TPT predictions. O-TPT and A-TPT show the same separation between range deviation and calibration quality without explicitly matching the zero-shot range. Exact logit-range alignment is thus insufficient to characterize the calibration of the full predictive distribution.

\subsection{Explaining Calibration Degradation via Zero-Shot-Relative Entropy}
\label{sec:entropy_analysis}

The previous analysis identifies predictive sharpening as a major source of calibration change and shows that logit range provides an incomplete description of the predictive distribution. We use predictive entropy to quantify the concentration of all class probabilities. For sample $x_i$, the normalized entropy change relative to its paired zero-shot prediction is
\begin{equation}
\delta_i^H
=
\frac{
H(\mathbf{p}_i^{\mathrm{TTA}})-H(\mathbf{p}_i^{\mathrm{ZS}})
}{
\log K
}.
\label{eq:relative_entropy}
\end{equation}
A negative $\delta_i^H$ indicates that adaptation produces a sharper predictive distribution than zero-shot CLIP.

\begin{figure*}[t]
    \centering
    \includegraphics[width=0.9\textwidth]{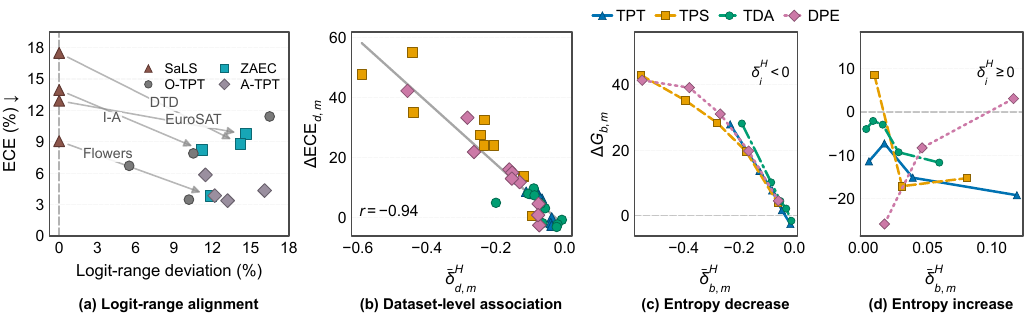}
    \caption{\textbf{Diagnostic analyses of TTA-induced calibration degradation.}
    (a) Exact zero-shot logit-range matching can retain high calibration error. Gray arrows connect SaLS and ZAEC applied to the same TPT predictions. (b) At the dataset--method level, stronger zero-shot-relative entropy reduction is associated with larger ECE degradation. (c) Among entropy-reduced samples, stronger entropy reduction corresponds to a larger confidence--accuracy gap. (d) Samples without entropy reduction show no consistent relationship.}
    \label{fig2}
    \vspace{-0.2cm}
\end{figure*}

\paragraph{Dataset--method-level association.}
Let $d$ and $m$ index the dataset and TTA method, respectively. We define the mean zero-shot-relative entropy change and the corresponding ECE change as
\begin{equation}
\begin{aligned}
\bar{\delta}_{d,m}^{H}
&=
\frac{1}{N_d}
\sum_{i=1}^{N_d}
\delta_{i,d,m}^{H},\\
\Delta\mathrm{ECE}_{d,m}
&=
\mathrm{ECE}_{d,m}^{\mathrm{TTA}}
-
\mathrm{ECE}_{d}^{\mathrm{ZS}}.
\end{aligned}
\label{eq:dataset_entropy_ece}
\end{equation}
Figure~\ref{fig2}(b) shows a strong association across datasets and methods: more negative $\bar{\delta}_{d,m}^{H}$ values coincide with larger ECE increases. The zero-shot-relative entropy change provides a label-free measure of the predictive concentration introduced by adaptation.

\paragraph{Sample-level confidence--accuracy gap.}
We further examine the relation within individual samples on ImageNet-A. For each method, samples are partitioned by the sign of $\delta_i^H$, and equal-frequency bins are constructed separately within each subset. For the $b$-th bin $B_{b,m}$, we define
\begin{equation}
\begin{aligned}
\bar{\delta}_{b,m}^{H}
&=
\frac{1}{|B_{b,m}|}
\sum_{i\in B_{b,m}}
\delta_i^H,\\
\Delta G_{b,m}
&=
\Delta\mathrm{Conf}_{b,m}
-
\Delta\mathrm{Acc}_{b,m}.
\end{aligned}
\label{eq:bin_entropy_gap}
\end{equation}
A positive $\Delta G_{b,m}$ means that confidence increases more than empirical accuracy. Figure~\ref{fig2}(c) shows a consistent trend within $\delta_i^H<0$: stronger entropy reduction corresponds to a larger increase in the confidence--accuracy gap for every evaluated method. Figure~\ref{fig2}(d) shows no consistent relation within $\delta_i^H\geq 0$. The observed calibration degradation is concentrated among predictions sharpened below their paired zero-shot entropy level.

\textbf{Observation 2: Zero-Shot-Relative Entropy Reduction Tracks Calibration Degradation.}
Across methods and datasets, larger entropy reductions relative to paired zero-shot predictions are associated with larger ECE increases. Within entropy-reduced samples, the confidence gain frequently exceeds the accuracy gain, indicating predictive sharpening unsupported by a corresponding improvement in correctness.

These findings identify zero-shot-relative entropy reduction as a sample-level, label-free signal of adaptation-induced predictive sharpening. They motivate a selective calibration procedure that acts on entropy-reduced predictions and uses each sample's zero-shot uncertainty as its reference.

%% file: sections/4-method1.tex
\section{Method}
\label{sec:method}
Motivated by the preceding analysis, we introduce Zero-Shot-Anchored Entropy Calibration (ZAEC), a label-free post-hoc procedure that selectively restores zero-shot entropy through sample-wise temperature scaling while preserving the adapted class ranking.
 
\subsection{Zero-Shot Entropy as a Sample-Wise Anchor}

Predictive uncertainty varies across samples. We use the
paired zero-shot entropy $H(\mathbf{p}_i^{\mathrm{ZS}})$ as
a sample-specific reference for measuring the concentration
introduced by adaptation. ZAEC corrects samples whose adapted
entropy falls below this reference:
\[
\mathcal{I}_{-}
=
\left\{
i \mid
H(\mathbf{p}_i^{\mathrm{TTA}})
<
H(\mathbf{p}_i^{\mathrm{ZS}})
\right\}.
\]
All remaining predictions are left unchanged.
\subsection{One-Sided Entropy Matching}
\label{sec:one_sided_entropy_matching}

For each adapted logit vector, define the entropy induced by temperature $T$ as
\begin{equation}
h_i(T)
=
H\left(
\operatorname{softmax}
\left(
\frac{\mathbf{z}_i^{\mathrm{TTA}}}{T}
\right)
\right),
\qquad T>0.
\label{eq:zaec_temperature_entropy}
\end{equation}
A temperature $T\geq 1$ smooths the adapted predictive distribution. For each sample in $\mathcal{I}_{-}$, ZAEC selects the smallest temperature that restores its paired zero-shot entropy. Samples outside $\mathcal{I}_{-}$ use the original adapted logits:
\begin{equation}
T_i^{\star}
=
\begin{cases}
\displaystyle
\min
\left\{
T\geq 1
\ \middle|\ 
h_i(T)
=
H\left(\mathbf{p}_i^{\mathrm{ZS}}\right)
\right\},
& i\in\mathcal{I}_{-},
\\[12pt]
1,
& i\notin\mathcal{I}_{-}.
\end{cases}
\label{eq:zaec_temperature}
\end{equation}
The calibrated prediction is
\begin{equation}
\widetilde{\mathbf{p}}_i
=
\operatorname{softmax}
\left(
\frac{\mathbf{z}_i^{\mathrm{TTA}}}{T_i^{\star}}
\right).
\label{eq:zaec_output}
\end{equation}
Equation~\eqref{eq:zaec_temperature} imposes a sample-specific entropy floor. It removes the entropy reduction introduced below the zero-shot reference and applies no correction to the other samples. ZAEC requires no target labels or learned calibration parameters. We compute $T_i^{\star}$ by bisection; the search bounds, stopping criterion, and numerical details are provided in the appendix.

\subsection{Conservative and Decision-Preserving Properties}
\label{sec:zaec_properties}

The following results characterize the effect of the one-sided temperature correction. Complete proofs are provided in the appendix.

\noindent\textbf{Proposition 1 (Monotone and Decision-Preserving Smoothing).}
For any non-constant logit vector $\mathbf{z}\in\mathbb{R}^{K}$, define
\begin{equation}
\mathbf{q}(T)
=
\operatorname{softmax}\left(\frac{\mathbf{z}}{T}\right),
\qquad T>0.
\label{eq:temperature_family}
\end{equation}
Then $H(\mathbf{q}(T))$ is continuous and strictly increasing in $T$, and the top-label confidence $\max_k q_k(T)$ is strictly decreasing in $T$. Positive temperature scaling preserves the complete class ranking: for any classes $a$ and $b$,
\begin{equation}
z_a>z_b
\quad\Longleftrightarrow\quad
q_a(T)>q_b(T).
\label{eq:temperature_ranking}
\end{equation}
Temperature thus provides a one-dimensional control of predictive concentration without changing the adapted decision.

\medskip
\noindent\textbf{Proposition 2 (Minimal Zero-Shot-Anchored Correction).}
Assume that $H(\mathbf{p}_i^{\mathrm{ZS}})<\log K$ for every $i\in\mathcal{I}_{-}$. Define the feasible temperature set
\begin{equation}
\mathcal{F}_i
=
\left\{
T\geq 1
\ \middle|\ 
H\left(
\operatorname{softmax}
\left(
\frac{\mathbf{z}_i^{\mathrm{TTA}}}{T}
\right)
\right)
\geq
H\left(\mathbf{p}_i^{\mathrm{ZS}}\right)
\right\}.
\label{eq:zaec_feasible_set}
\end{equation}
For $i\in\mathcal{I}_{-}$, ZAEC selects the unique minimum feasible temperature,
\begin{equation}
T_i^{\star}=\min\mathcal{F}_i,
\label{eq:zaec_minimum_temperature}
\end{equation}
and its output satisfies
\begin{equation}
H\left(\widetilde{\mathbf{p}}_i\right)
=
\max\left\{
H\left(\mathbf{p}_i^{\mathrm{TTA}}\right),
H\left(\mathbf{p}_i^{\mathrm{ZS}}\right)
\right\}.
\label{eq:zaec_entropy_floor}
\end{equation}
Within the temperature-scaling family, $\widetilde{\mathbf{p}}_i$ is the closest feasible prediction to the raw TTA output in forward KL divergence:
\begin{equation}
T_i^{\star}
=
\underset{T\in\mathcal{F}_i}{\arg\min}\;
D_{\mathrm{KL}}
\left(
\mathbf{p}_i^{\mathrm{TTA}}
\,\middle\|\,
\operatorname{softmax}
\left(
\frac{\mathbf{z}_i^{\mathrm{TTA}}}{T}
\right)
\right).
\label{eq:zaec_kl_minimality}
\end{equation}
The complete class ranking is preserved:
\begin{equation}
\operatorname{argsort}\left(\widetilde{\mathbf{p}}_i\right)
=
\operatorname{argsort}\left(\mathbf{p}_i^{\mathrm{TTA}}\right).
\label{eq:zaec_argsort}
\end{equation}
For samples outside $\mathcal{I}_{-}$, $T_i^{\star}=1$ and the adapted prediction remains unchanged. For samples in $\mathcal{I}_{-}$, ZAEC applies the minimum smoothing required to reach the sample-specific entropy floor. The procedure preserves the top-1 prediction and classification accuracy for every sample.

%% file: sections/5-experiment.tex
\begin{figure}[h]
    \centering
    \includegraphics[width=0.9\linewidth]{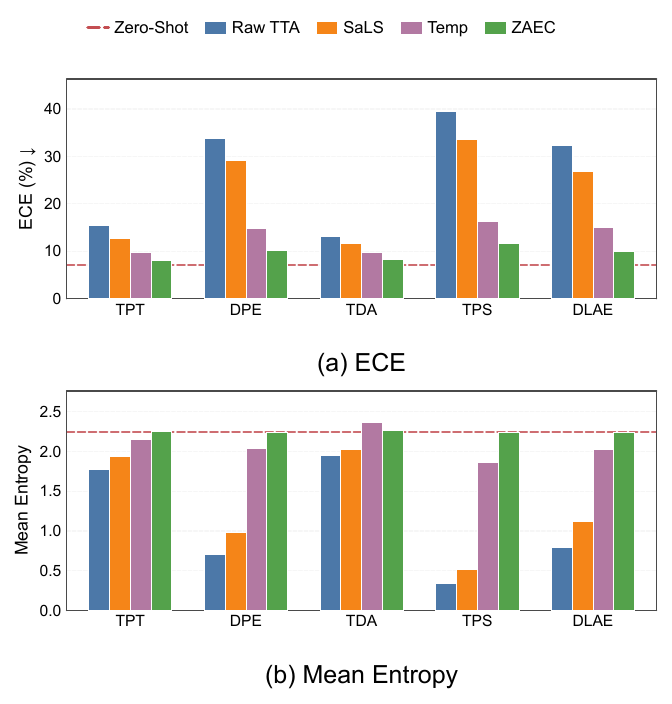}
    \caption{\textbf{Average ECE and predictive entropy on four ImageNet natural distribution-shift datasets using CLIP RN50.}
    ZAEC restores mean predictive entropy toward the zero-shot reference and achieves the lowest average ECE for each evaluated TTA method.}
    \label{fig:rn50_results}
\end{figure}
\section{Experiments}

\begin{table}[t]
\renewcommand{\arraystretch}{0.85}
\setlength{\tabcolsep}{3.5pt}
\small
\begin{tabular}{l|l|cccccc}
\toprule
\textbf{Method} &
\textbf{Metric} &
\textbf{I-A} &
\textbf{I-V} &
\textbf{I-R} &
\textbf{I-S} &
\textbf{Avg.} \\
\midrule

CLIP-ViT-B/16
& Acc.
 & 47.83 & 60.86 & 73.97 & 46.10 & 57.19 \\
& ECE
& 8.33 & 3.14 & 3.54 & 4.87 & 4.97 \\
\midrule

TPT
& Acc.
 & 52.01 & 62.69 & 76.54 & 47.96 & 59.80 \\
\quad + None
& ECE
 & 16.74 & 11.58 & 4.43 & 15.60 & 12.09 \\
\quad + Temp.Scal
& ECE
 & 10.52 & 6.31 & \textbf{1.62} & 8.50 & 6.74 \\
\quad + SaLS
& ECE
 & 13.95 & 9.67 & 2.49 & 13.77 & 9.97 \\
\quad + Ours
& ECE
& \textbf{8.19} & \textbf{4.21} & 3.24 & \textbf{6.92} & \textbf{5.64} \\
\midrule

TPS
& Acc.
 & 58.27 & 64.27 & 79.44 & 50.93 & 63.23 \\
\quad + None
& ECE
& 34.29 & 28.50 & 16.28 & 34.80 & 28.47 \\
\quad + Temp.Scal
& ECE
& 20.39 & 14.62 & 6.83 & 14.70 & 14.14 \\
\quad + SaLS
& ECE
 & 29.62 & 26.23 & 13.73 & 30.80 & 25.10 \\
\quad + Ours
& ECE
 & \textbf{9.03} & \textbf{10.18} & \textbf{1.59} & \textbf{11.95} & \textbf{8.19} \\
\midrule

TDA
& Acc.
& 49.03 & 61.28 & 75.01 & 48.73 & 58.51 \\
\quad + None
& ECE
& 15.21 & 15.09 & \textbf{1.50} & 11.92 & 10.93 \\
\quad + Temp.Scal
& ECE
& \textbf{8.27} & \textbf{4.01}  & 10.95 & \textbf{4.97} & 7.05 \\
\quad + SaLS
& ECE
& 13.31 & 12.57 & 2.05 & 9.39 & 9.33 \\
\quad + Ours
& ECE
 & 8.83 & 4.03 & 4.48 & 5.01 & \textbf{5.59} \\
\midrule

DPE
& Acc.
 & 58.05 & 65.18 & 79.69 & 52.54 & 63.87 \\
\quad + None
& ECE
 & 35.26 & 22.43 & 8.15 & 23.64 & 22.37 \\
\quad + Temp.Scal
& ECE
& 23.74 & 8.92 & 5.59 & 4.94 & 10.80 \\
\quad + SaLS
& ECE
& 31.74 & 18.02 & \textbf{4.79} & 17.80 & 18.09 \\
\quad + Ours
& ECE
& \textbf{11.09} & \textbf{5.62} & 5.09 & \textbf{4.91} & \textbf{6.68} \\
\midrule

DLAE
& Acc.
 & 58.95 & 65.29 & 79.75 & 52.37 & 64.09 \\
\quad + None
& ECE
 & 34.13 & 21.70 & 7.47 & 22.56 & 21.47 \\
\quad + Temp.Scal
& ECE
& 21.52 & 8.11 & 7.34 & \textbf{4.06} & 10.26 \\
\quad + SaLS
& ECE
& 30.00 & 16.81 & \textbf{3.77} & 16.19 & 16.69 \\
\quad + Ours
& ECE
 & \textbf{11.09} & \textbf{5.61} & 5.22 & 4.77 & \textbf{6.67} \\
\bottomrule
\end{tabular}
\caption{\textbf{Calibration performance across ImageNet natural distribution shift datasets with CLIP ViT-B/16.} The best-performing result within each TTA block is shown in \textbf{bold}. Avg. denotes the average over the four ImageNet variants.}
\label{tab:natural_shift}
\end{table}

\begin{table*}[t]
\centering
\renewcommand{\arraystretch}{0.85}
\setlength{\tabcolsep}{3.5pt}
\small
\begin{tabular}{l|l|cccccccccccc}
\toprule
\textbf{Method} &
\textbf{Metric} &
\textbf{INet}&
\textbf{DTD} &
\textbf{FLOW} &
\textbf{FOOD} &
\textbf{SUN} &
\textbf{EURO} &
\textbf{AIR} &
\textbf{PETS} &
\textbf{CALT} &
\textbf{UCF} &
\textbf{CAR} &
\textbf{Avg.} \\
\midrule

CLIP-ViT-B/16
& Acc.
&66.74 & 44.33 & 67.40 & 85.24 & 62.56 & 42.01 & 23.88
& 88.20 & 93.31 & 65.05 & 65.58 & 64.03 \\
& ECE
&1.96 & 8.23 & 2.82 & 2.15 & 2.19 & 7.22 & 5.42
& 4.45 & 5.07 & 3.06 & 4.45 & 4.27 \\
\midrule

TPT
& Acc.
&68.85 & 46.93 & 68.66 & 86.34 & 65.51 & 42.31 & 24.48
& 88.14 & 93.39 & 68.17 & 67.08 & 65.44 \\
\quad + None
& ECE
&9.64 & 21.12 & 10.87 & 2.29 & 11.45 & 15.96 & 11.86
& 2.73 & 3.56 & 11.08 & 1.87 & 9.31 \\
\quad + Temp.Scal
& ECE
& 5.17& 14.90 & 6.28 & 1.19 & 6.54 & 12.08 & 7.52
& 2.25 & 3.39 & 6.55 & 3.52 & 6.31 \\
\quad + SaLS
& ECE
&7.86& 17.50 & 9.06 & \textbf{1.12} & 9.37 & 12.92 & 10.78
& \textbf{2.15} & \textbf{2.91} & 8.76 & \textbf{1.29} & 7.61 \\
\quad + Ours
& ECE
&\textbf{3.22} & \textbf{8.79} & \textbf{3.82} & 1.57 & \textbf{3.56} & \textbf{9.73} & \textbf{6.16}
& 3.54 & 4.29 & \textbf{2.88} & 3.80 & \textbf{4.67} \\
\midrule

TPS
& Acc.
&70.44 & 52.31 & 72.72 & 83.87 & 63.86 & 44.56 & 26.40
& 81.52 & 93.67 & 68.70 & 68.75 & 66.07 \\
\quad + None
& ECE
&23.64 & 32.97 & 20.82 & 13.77 & 29.57 & 48.99 & 28.44
& 16.44 & 5.84 & 25.66 & 21.04 & 24.29 \\
\quad + Temp.Scal
& ECE
&12.05& 15.48 & 10.31 & 6.69 & 11.35 & 33.87 & 14.52
& 11.26 & \textbf{1.63} & 11.28 & 11.99 & 12.77 \\
\quad + SaLS
& ECE
&21.89 & 25.43 & 18.73 & 12.34 & 27.29 & 31.37 & 21.78
& 15.70 & 5.32 & 23.38 & 19.29 & 20.23 \\
\quad + Ours
& ECE
&\textbf{7.72} & \textbf{6.50} & \textbf{4.22} & \textbf{4.68} & \textbf{11.04} & \textbf{16.16} & \textbf{6.02}
& \textbf{8.76} & 3.39 & \textbf{7.72} & \textbf{5.05} & \textbf{7.39} \\
\midrule

TDA
& Acc.
&68.42 & 47.40 & 70.81 & 85.66 & 65.73 & 55.16 & 23.79
& 88.50 & 93.59 & 64.46 & 66.91 & 66.40 \\
\quad + None
& ECE
&6.12 & 16.50 & 8.87 & 0.94 & 7.06 & \textbf{6.30} & 17.97
& \textbf{2.35} & 3.32 & 8.88 & 3.83 & 7.47 \\
\quad + Temp.Scal
& ECE
&4.46 & 8.93 & 2.89 & 6.78 & 4.11 & 6.76 & 9.18
& 5.57 & 3.82 & 5.59 & 6.47 & 5.87 \\
\quad + SaLS
& ECE
& 4.62& 12.44 & 6.65 & \textbf{0.92} & 5.38 & 6.98 & 14.37
& 3.29 & \textbf{3.09} & 7.21 & \textbf{2.58} & 6.14 \\
\quad + Ours
& ECE
&\textbf{2.05}& \textbf{7.77} & \textbf{2.86} & 2.25 & \textbf{2.54} & 7.11 & \textbf{8.71}
& 4.43 & 4.22 & \textbf{2.80} & 3.20 & \textbf{4.36} \\
\midrule

DPE
& Acc.
&71.57 & 55.32 & 74.50 & 86.78 & 70.18 & 49.37 & 30.36
& 92.12 & 95.74 & 73.80 & 69.13 & 69.90 \\
\quad + None
& ECE
&16.21 & 24.72 & 16.64 & 6.98 & 17.30 & 37.01 & 34.42
& 3.43 & 1.69 & 13.94 & 17.35 & 17.24 \\
\quad + Temp.Scal
& ECE
&4.30 & 8.32 & 6.05 & \textbf{1.26} & 4.23 & 26.20 & 14.70
& 3.99 & 3.11 & 3.23 & 4.23 & 7.24 \\
\quad + SaLS
& ECE
&12.01 & 15.09 & 12.93 & 4.55 & 12.59 & 17.12 & 24.96
& \textbf{1.91} & \textbf{0.92} & 10.69 & 14.98 & 11.61 \\
\quad + Ours
& ECE
&\textbf{2.30}& \textbf{6.35} & \textbf{3.80} & 1.49 & \textbf{2.11} & \textbf{11.23} & \textbf{7.49}
& 4.36 & 5.38 & \textbf{2.47} & \textbf{2.50} & \textbf{4.50} \\
\midrule

DLAE
& Acc.
&71.47 & 55.20 & 74.83 & 86.82 & 70.42 & 58.30 & 30.15
& 92.80 & 96.02 & 73.72 & 70.29 & 70.91 \\
\quad + None
& ECE
&15.68& 22.28 & 14.66 & 6.71 & 16.41 & 18.15 & 23.13
& 2.91 & 1.79 & 12.64 & 13.35 & 13.43 \\
\quad + Temp.Scal
& ECE
&4.17& 6.96 & 4.55 & 2.04 & 3.50 & \textbf{3.67} & 5.72
& 5.48 & 3.84 & 3.58 & 2.32 & 4.17 \\
\quad + SaLS
& ECE
&11.10& 13.39 & 10.28 & 3.99 & 11.22 & 4.26 & 14.95
& \textbf{1.74} & \textbf{1.43} & 8.64 & 10.79 & 8.34 \\
\quad + Ours
& ECE
&\textbf{2.38}& \textbf{6.54} & \textbf{3.69} & \textbf{1.33} & \textbf{1.92} & 5.57 & \textbf{5.10}
& 4.78 & 5.45 & \textbf{3.22} & \textbf{2.22} & \textbf{3.84} \\

\bottomrule
\end{tabular}
\caption{Calibration performance across ImageNet and fine-grained classification datasets with CLIP ViT-B/16. The best-performing result within each TTA block is shown in \textbf{bold}. Avg. denotes the average over all 11 datasets.}
\label{tab:fine_grained}
\end{table*}

\subsection{Experimental Setup}

\paragraph{Datasets.}
We evaluate calibration performance under two complementary settings.
The natural distribution-shift setting contains four ImageNet variants:
ImageNet-A~\cite{inet-a}, ImageNet-V2~\cite{inet-v2},
ImageNet-R~\cite{inet-r}, and ImageNet-Sketch~\cite{inet-k}.
The second setting contains ImageNet~\cite{imagenet} and ten diverse
classification datasets commonly used in prior vision-language model
adaptation studies: DTD~\cite{dtd}, OxfordFlowers~\cite{flowers102},
Food101~\cite{food101}, SUN397~\cite{sun},
EuroSAT~\cite{eurosat}, Aircraft~\cite{aircraft},
OxfordPets~\cite{pets}, Caltech101~\cite{caltech101},
UCF101~\cite{ucf101}, and StanfordCars~\cite{stanfordcars}.

\paragraph{Baselines.}
We evaluate five representative VLM TTA methods: TPT~\cite{tpt}, TPS~\cite{tps}, TDA~\cite{tda}, DPE~\cite{dpe}, and DLAE~\cite{dlae}. Together, they cover prompt optimization, prototype adaptation, dynamic caches, and logit adjustment. For each method, we compare the raw adapted output with SaLS~\cite{sals}, Temperature Scaling~\cite{guo2017calibration}, and ZAEC. 
\begin{figure}
    \centering
    \includegraphics[width=1.0\linewidth]{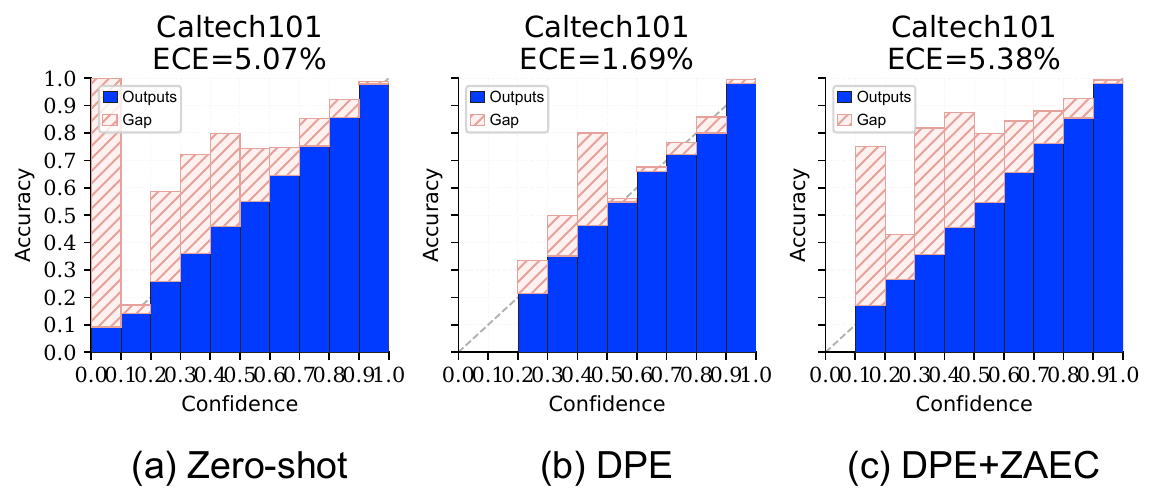}
    \caption{\textbf{Boundary case of ZAEC on Caltech101.}
       }
    \label{fig:caltech101}
        \vspace{-0.3cm}
\end{figure}
\paragraph{Implementation details.}We evaluate CLIP with ViT-B/16~\cite{vit} and RN50~\cite{resnet}, reporting ViT-B/16 results in the main paper and complete RN50 results in the appendix. Experiments follow the TTABC benchmark~\cite{huang2026drives}, with each TTA method reproduced using its original configuration. All experiments were conducted on a single Tesla V100S-PCIE-32GB GPU.

\paragraph{Evaluation metrics.}
We report top-1 accuracy and Expected Calibration Error. We additionally report negative log-likelihood (NLL) in the  ablation. All dataset-group averages are unweighted macro-averages over datasets. Reliability diagrams use 10 bins for visualization only.

\subsection{Main Results}
\begin{figure}
    \centering
    \includegraphics[width=0.9\linewidth]{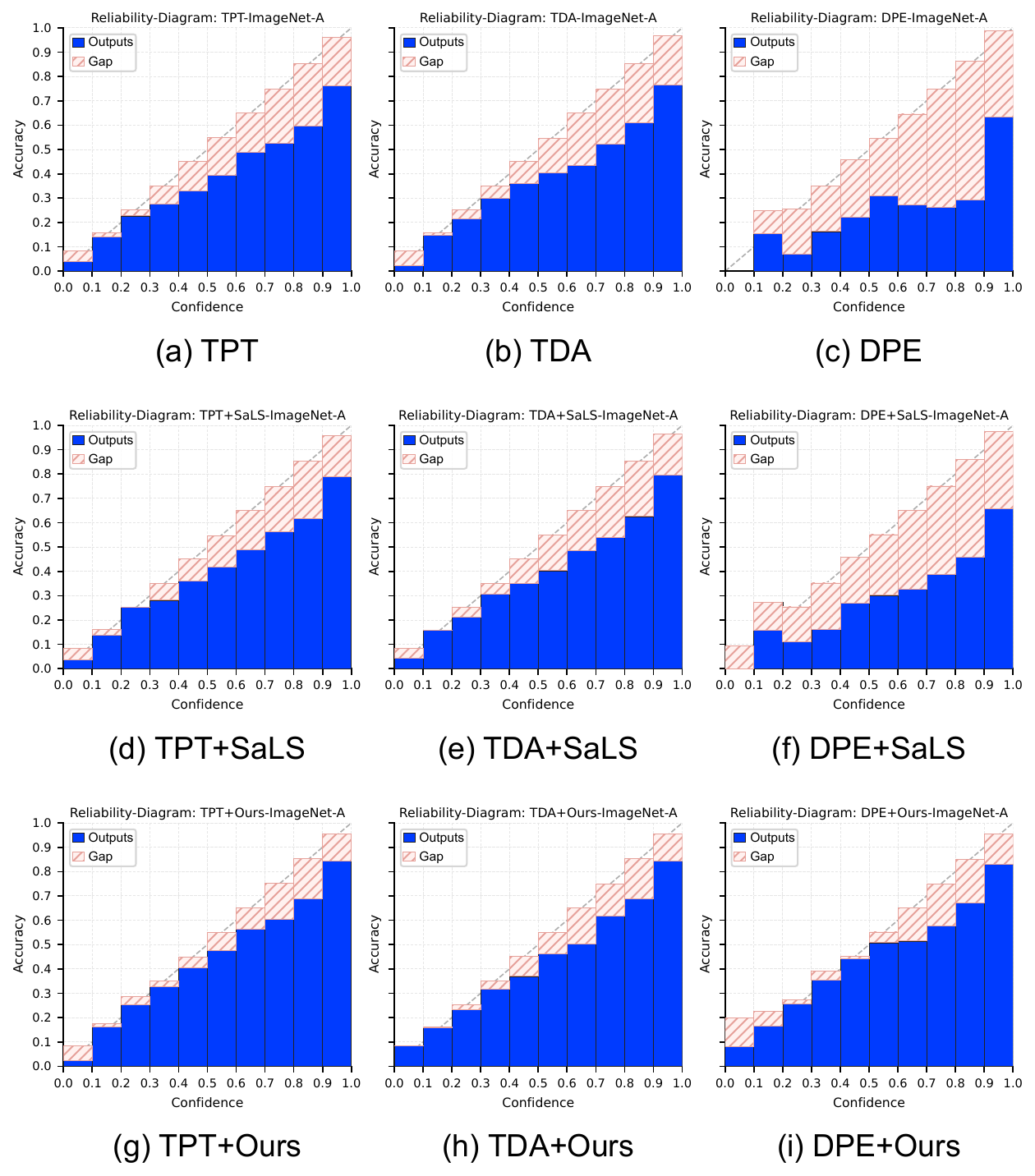}
    \caption{\textbf{Reliability diagrams on ImageNet-A using CLIP ViT-B/16.} Rows correspond to raw TTA, TTA+SaLS, and TTA+ZAEC, while columns correspond to TPT, TDA, and DPE. Predictions are grouped into 10 confidence bins. Blue bars show empirical bin accuracy, and hatched regions indicate the confidence–accuracy gap.}
    \label{fig:reliability}
\end{figure}
Table~\ref{tab:natural_shift} and Table~\ref{tab:fine_grained} show that, although the evaluated TTA methods generally improve classification accuracy over zero-shot CLIP, their raw outputs often suffer from substantial calibration degradation. Across both benchmark groups, the raw TTA methods yield method-level macro-average ECE values ranging from 7.47\% to 28.47\%. ZAEC reduces these values to a range of 3.84\% to
8.19\% while preserving class rankings and classification accuracy. It achieves the lowest method-level average ECE among the evaluated
post-hoc methods for every TTA method and benchmark group, including source-fitted Temperature Scaling. Results with standard deviations are provided in Appendix D.
Figure ~\ref{fig:rn50_results} reports average ECE and predictive entropy on the RN50 backbone. On CLIP RN50, raw TTA lowers mean predictive entropy for several adaptation methods. ZAEC raises entropy toward the zero-shot level and achieves the lowest method-level average ECE across the evaluated calibration approaches. The RN50 results reproduce the pattern observed with ViT-B/16 across a second backbone.

The reliability diagrams in Figure~\ref{fig:reliability} further show that ZAEC generally reduces the confidence--accuracy gaps of TPT, TDA, and DPE on ImageNet-A, while residual miscalibration remains in several confidence bins.

\subsection{Ablation Study}
\textbf{Ablation of the Corrected Subset.}
We examine which adapted predictions should be calibrated.
\textit{All predictions} matches every adapted prediction to
the entropy of its paired zero-shot prediction.
\textit{Entropy-increased} applies matching only when
$H(\mathbf{p}_i^{\mathrm{TTA}})
>
H(\mathbf{p}_i^{\mathrm{ZS}})$,
which sharpens the adapted distribution.
\textit{Entropy-reduced} applies matching only when
$H(\mathbf{p}_i^{\mathrm{TTA}})
<
H(\mathbf{p}_i^{\mathrm{ZS}})$
and corresponds to the proposed ZAEC.
All settings use the same sample-wise temperature-scaling
procedure.

\begin{center}
\renewcommand{\arraystretch}{0.9}
\setlength{\tabcolsep}{4.5pt}
\small
\begin{tabular}{lcccc}
\toprule
Corrected subset
& \multicolumn{2}{c}{TPT}
& \multicolumn{2}{c}{TDA} \\
\cmidrule(lr){2-3}\cmidrule(lr){4-5}
& ECE & NLL & ECE & NLL \\
\midrule
None (Raw TTA)
& 11.60 & 1.67 & 9.97 & 1.65 \\
All predictions
& 5.29 & 1.61 & 5.01 & 1.63 \\
Entropy-increased
& 11.82 & 1.68 & 9.31 & 1.67 \\
Entropy-reduced (ZAEC)
& \textbf{5.16} & \textbf{1.60}
& \textbf{4.88} & \textbf{1.61} \\
\bottomrule
\end{tabular}
\captionof{table}{
\textbf{Ablation of the corrected subset on the five
ImageNet-family benchmarks.}
ECE and NLL are macro-averaged over the five datasets.
NLL denotes negative log-likelihood.
}
\label{tab:ZAEC_mode_ablation}
\end{center}

Table~\ref{tab:ZAEC_mode_ablation} shows that correcting
entropy-increased predictions provides little calibration
benefit. Matching all predictions substantially reduces ECE,
and restricting the correction to entropy-reduced predictions
achieves the lowest ECE and NLL for both TPT and TDA.
The gains are thus concentrated among predictions sharpened
by adaptation, supporting the one-sided design of ZAEC.

\begin{figure}[t]
    \centering
    \includegraphics[width=1.0\linewidth]{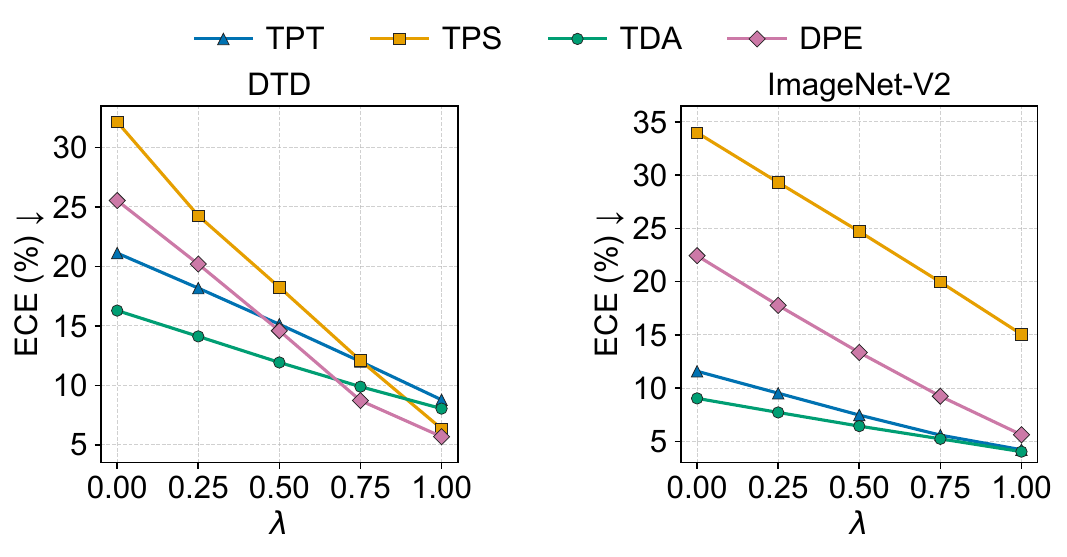}
    \caption{
    \textbf{Ablation of entropy-restoration strength on DTD and ImageNet-V2 with CLIP ViT-B/16.}
    }
    \label{fig:entropy_strength}
    \vspace{-0.3cm}
\end{figure}

\textbf{Ablation of Entropy-Matching Strength.}
We next examine how the amount of entropy restoration
affects calibration. For each entropy-reduced sample
$i$ satisfying
$H(\mathbf p_i^{\mathrm{TTA}})
< H(\mathbf p_i^{\mathrm{ZS}})$,
we define the target entropy as
\begin{equation}
H_i^{\mathrm{target}}(\lambda)
=
H\!\left(\mathbf{p}_i^{\mathrm{TTA}}\right)
+
\lambda
\left[
H\!\left(\mathbf{p}_i^{\mathrm{ZS}}\right)
-
H\!\left(\mathbf{p}_i^{\mathrm{TTA}}\right)
\right],
\label{eq:partial_restoration}
\end{equation}
where
$\lambda \in \{0, 0.25, 0.5, 0.75, 1\}$.
The corresponding temperature is obtained using the same
bisection procedure as ZAEC, while samples outside
$\mathcal{I}_{-}$ remain unchanged.

Here, $\lambda=0$ recovers raw TTA and $\lambda=1$ recovers
ZAEC; all variants preserve the adapted class ranking.

Figure~\ref{fig:entropy_strength} shows that ECE decreases
monotonically with $\lambda$ for all evaluated methods on both DTD and ImageNet-V2, with full restoration
achieving the lowest ECE.
This result supports the selected restoration target in the
evaluated setting, rather than establishing its universal
optimality across all datasets and adaptation methods.

\textbf{Boundary case on Caltech101.} Figure \ref{fig:caltech101} shows a boundary case where the zero-shot uncertainty is an overly conservative reference. Zero-shot CLIP is underconfident across most confidence bins on Caltech101, with empirical accuracy consistently exceeding predicted confidence. DPE raises confidence and brings the reliability profile closer to the diagonal, reducing ECE from 5.07\% to 1.69\%. ZAEC restores the predictive entropy toward the zero-shot level and lowers this useful confidence, causing the model to become underconfident again and increasing ECE to 5.38\%. This case indicates that ZAEC can be less effective when TTA-induced sharpening already corrects zero-shot underconfidence.

%% file: sections/6-conclusion.tex



\section{Conclusion}

We studied calibration degradation in test-time-adapted VLMs and identified prediction-preserving sharpening as an important failure mode, with zero-shot-relative entropy reduction providing a label-free signal of this behavior. Based on this finding, we introduced ZAEC, a selective post-hoc method that restores zero-shot entropy through minimal sample-wise temperature scaling without changing class rankings or classification accuracy. Across five TTA methods and 15 datasets, ZAEC achieves the lowest
method-level average ECE among the evaluated post-hoc methods on
ViT-B/16, with consistent gains on RN50. Future work will explore adaptive anchors for cases where zero-shot underconfidence makes ZAEC overly conservative.